\documentclass[conference]{IEEEtran}
\usepackage{cite}
\usepackage{amsmath,amssymb,amsfonts}
\usepackage{algorithmic}
\usepackage{graphicx}
\usepackage{textcomp}
\usepackage{xcolor}
\usepackage{url}
\def\BibTeX{{\rm B\kern-.05em{\sc i\kern-.025em b}\kern-.08em
    T\kern-.1667em\lower.7ex\hbox{E}\kern-.125emX}}
\begin{document}

\title{Veni, Vidi, Vici: Solving the Myriad of Challenges before Knowledge Graph Learning}

\author{\IEEEauthorblockN{1\textsuperscript{st} Jeffrey Sardina}
\IEEEauthorblockA{\textit{School of Computer Science and Statistics} \\
\textit{Trinity College Dublin and Accenture Labs}\\
Dublin, Ireland \\
0000-0003-0654-2938}
\and
\IEEEauthorblockN{2\textsuperscript{nd} Luca Costabello}
\IEEEauthorblockA{\textit{BioInnovation Labs} \\
\textit{Accenture}\\
Dublin, Ireland \\
0000-0002-0720-9347}
\and
\IEEEauthorblockN{3\textsuperscript{rd} Christophe Guéret}
\IEEEauthorblockA{\textit{BioInnovation Labs} \\
\textit{Accenture}\\
Dublin, Ireland \\
0000-0002-8914-6107}
}

\maketitle

\begin{abstract}
Knowledge Graphs (KGs) have become increasingly common for representing large-scale linked data. However, their immense size has required graph learning systems to assist humans in analysis, interpretation, and pattern detection. While there have been promising results for researcher- and clinician- empowerment through a variety of KG learning systems, we identify four key deficiencies in state-of-the-art graph learning that simultaneously limit KG learning performance and diminish the ability of humans to interface optimally with these learning systems. These deficiencies are: 1) lack of expert knowledge integration, 2) instability to node degree extremity in the KG, 3) lack of consideration for uncertainty and relevance while learning, and 4) lack of explainability. Furthermore, we characterise state-of-the-art attempts to solve each of these problems and note that each attempt has largely been isolated from attempts to solve the other problems. Through a formalisation of these problems and a review of the literature that addresses them, we adopt the position that not only are deficiencies in these four key areas holding back human-KG empowerment, but that the divide-and-conquer approach to solving these problems as individual units rather than a whole is a significant barrier to the interface between humans and KG learning systems. We propose that it is only through integrated, holistic solutions to the limitations of KG learning systems that human and KG learning co-empowerment will be efficiently affected. We finally present our "Veni, Vidi, Vici" framework that sets a roadmap for effectively and efficiently shifting to a holistic co-empowerment model in both the KG learning and the broader machine learning domain.
\end{abstract}


\section{Introduction}
Knowledge Graphs (KGs) are semantic data stores that model data as a set of nodes and the relations between them \cite{kg-ovewview}. The atomic unit of knowledge in a knowledge graph is the triple, which consists of a single labelled head node, a labelled tail node, and a directed, labelled edge that relates the head to the tail. For example, the fact "Sauron created the One Ring" could be written as \textit{(Sauron, created, One-Ring)}.

The graph structure of KGs lends them very naturally to a variety of intrinsically networked datasets, such as social networks, computer networked systems, and biomedical drug-gene interaction networks. However, KG data has grown to a size that precludes human analysis without computational assistance. For example, Hetionet, a common benchmark dataset for drug repurposing in bioinformatics, contains 2.25M triples -- far more than a human can analyse alone \cite{hetionet}. Other large KGs can range from 1M to 21M triples \cite{light-into-the-dark,conceptnet}.

In order to empower human use of these massive KGs, various KG learning systems have been proposed to detect patterns and predict new information in the domain of a given KG. Knowledge Graph Embeddings (KGEs) use machine learning to attempt to represent the semantics of a KG in vector space \cite{kg-ovewview,rml-review,kge-survey}. Other KG learning approaches include logical and rule-based methods \cite{PoLo,dl-learner,kge-logic-survey}, path-based methods \cite{PoLo,rdf2vec}, and Graph Neural Networks (GNNs) \cite{sl-dsgcn,gnn-review,vgae,CensNet}.

KG learning systems have made significant progress in assisting with research tasks, such as drug re-purposing \cite{PoLo,kg-filtering,kg-filtering} and drug-drug interaction prediction \cite{kges-for-ddi,kges-for-ddi-2}. However, recent literature has identified some major shortcomings of the current state-of-the-art approaches to KG learning systems.

The first shortcoming is that most existing KG learning systems do not account for the logical structure of the data, which means that these systems typically cannot use an expert-curated ontologies even if they are available \cite{kg-ovewview,rml-review,kge-survey,gnn-review,onto-in-neg-sampler}. Moreover, application of KGEs to real-world problems, notably including high-potential-impact uses such as medical drug repurposing and predictions about drugs and diseases in the biomedical context, are mostly done on methods that make no attempt to model the logical structure of the data \cite{realistic-ddi,kge-rare-disease,discover-protein-targets,target-drug-discovery}. This means that many logical and causal inferences the humans care about most are not well accounted for, either in theory or in practice.

Moreover, KG learning systems have been shown to suffer from biases due to varying distributions of node degrees: high-degree nodes are generally learned much more reliably and with much greater embedding quality than low-degree nodes across various KG learning methods \cite{topological-imbalance,meta-learning-tail,meta-tail-gnn}. Despite their generally superior embedding quality, high-degree nodes can also have detrimental effects on KG learning systems \cite{kg-filtering,topological-imbalance}, to the point that removing high-degree domain-relevant data from the KG can boost performance \cite{kg-filtering}.

KG learning systems also often lack the focus that is desired by human researchers and clinicians: while in many cases KG learning systems are only used for predicting certain types of facts in a graph, those models are trained to predict all types of facts with equal emphasis on each one \cite{kg-filtering}. Most widely-used KG learning systems do not calculate or use uncertainty or relevance scores for triples, which further prevents the model from focusing on the most relevant and most reliable information \cite{kg-ovewview,rml-review,kge-survey,focuse}.

Finally, most KG learning systems were not created or evaluated for explainability \cite{rml-review,kge-survey,meta-tail-gnn,PoLo,rdf2vec,gnn-review,meta-learning-tail,explaining-kges}. Some logical methods do allow explainability \cite{PoLo,dl-learner,amie}; however, with few exceptions, these logical approaches generally cannot handle large KGs \cite{PoLo}. Moreover, there is no agreed-upon definition or concrete metric for explainability for KG learning systems, which makes comparison of progress in this area a challenge in itself \cite{explaining-kges,xai-metrics}. As a result, not only are KG learning systems opaque to humans, but critically \textit{detecting the presence and cause of the other 3 problems is severely hindered by the black-box nature of most KG learning systems}.

Individually, these problems are each a major setback for human-AI empowerment and advancements in the realm of KGs. Taken together, they present a bleak outlook for the use of KG learning techniques and suggest an urgent need for corrective innovation and development within the field.

In this position paper, we first formalise definitions and nomenclatures for each of these four problems: \textit{lack of expert knowledge integration}, \textit{instability to extreme topological variation}, \textit{inability to perform focused learning}, and \textit{lack of explainability}. We present the details of each problem and give a review of state-of-the-art techniques that attempt to address each one. Our analysis leads us to the conclusion that the state-of-the-art is deficient not only because of a general lack of robust answers to these problems, but because existing solutions tend to address only one of these problems while ignoring the others. We conclude by presenting our "Veni, Vidi, Vici" framework for how to represent and consider diverse problems in the same conceptual space to best empower humans with robust, interpretable, and reliable learning systems. Finally, we propose that use of the "Veni, Vidi, Vici" model will allow researchers to not only greatly advance KG learning systems past these major challenges, but also fundamentally strengthen the machine learning-human interface in many different domains.

\section{Expert Knowledge Integration}
We first define expert knowledge as so: \textit{expert knowledge is knowledge about the logic, dependency patterns, pragmatics, inferences, and analytic approaches of a specific domain}. From this, we formally define expert knowledge integration as follows: \textit{expert knowledge integration is the act of making a machine learning model explicitly aware of expert knowledge in how it models tasks and/or data}. While in different domains this will have widely different manifestations, in the domain of KG learning systems it can be presented in a very straight forward manner. We can say, without loss of generality, that \textit{a KG learning system expresses expert knowledge integration if and only if it is able to recognise different information with the exact same graph topology.}

\begin{figure}
\centering
\includegraphics[width=3in]{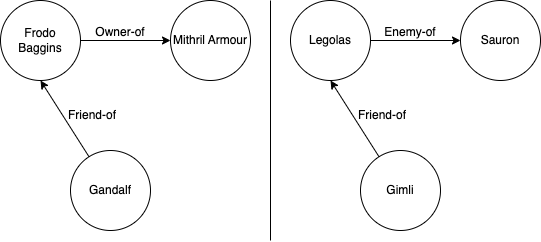}
\caption{Two graphs with identical topology but distinct information.} \label{fig1}
\end{figure}

For example, take Figure~\ref{fig1}. To a human, it is obvious that if Gimli is a friend of Legolas and Legolas is an enemy of Sauron, that Gimli should also be an enemy of Sauron. On the other hand, there is no reason to assume a direct connection between Gandalf and Mithril Armour based on the information given in the left-hand graph. The two graphs are topologically identical, but very distinct in information content.

Modern KGEs, such as TransE, ComplEx, DistMult, RotatE, and ConvE \cite{rml-review,kge-survey,kge-logic-survey}, as well as rule mining systems such as AMIE+ \cite{amie}, GNNs \cite{gnn-review,vgae,CensNet}, and path-based KG learning systems \cite{rdf2vec} cannot take this expert information into account and would learn these two graphs identically. Using this example as a guide, we formalise the problem of expert knowledge integration as so: \textit{By what methods is it possible to teach a KG learning system to distinguish knowledge with identical topology but variable semantics?} We call methods that attempt to distinguish such graphs "Expert Knowledge Methods".

Note: although rule-mining and query-based approaches are at times discussed as if they contained expert knowledge, rule-based systems such as AMIE+ and IterE \cite{amie,itere} and query-answering methods that allow logical operations in queries (such as Query2box and CQD) \cite{Query2box,cqd} are not Expert Knowledge Systems. We exclude them because none of them can distinguish the two graphs in Figure~\ref{fig1}; i.e., their reasoning derives entirely from topological patterns, not from domain expertise or expert knowledge.

\subsection{Expert Knowledge Methods}
Modern literature has principally focused on two main methods for creating expert-knowledge aware systems: use of ontological rules \cite{dl-learner,kge-logic-survey,onto-in-neg-sampler,deductive-completion,e2r,el-embeddings} and use of non-ontological domain expertise \cite{PoLo,kg-filtering}.

Ontology-based methods incorporate background knowledge expressed as ontological rules at various stages of the learning pipeline. For example, Alshahrani et al. use an ontology to complete an existing KG before attempting to learn on it, meaning that all rules inferred by the ontology are explicitly present in the graph \cite{deductive-completion}. In contrast, DL-Learner mines rules from an KG and uses them to refine or expand an existing ontology, taking the initial form of that ontology into account \cite{dl-learner}. Injecting ontologies into training is also done, either to create negative examples to learn from \cite{onto-in-neg-sampler} or to  explicitly model for logical relationships during training \cite{e2r,el-embeddings}. A more complete analysis of the stages of ontology integration and ontology-guided learning can be found in \cite{kge-logic-survey}. 

A second, more recent approach to integrating expert knowledge is using "meta-paths" procured by domain experts to guide learning along specific parts of a KG. Meta-paths are paths of entities and relationships in a KG that are identified not by the identity of nodes along them, but by their broader type. For example, \textit{(Gandalf, friends-with, Aragorn)} would have the meta-path \textit{(Person, friends-with, Person)}. These meta-paths allow expects to create rules that graph learners can use to bias their training. For example, PoLo uses expert-curated meta-paths to guide a path-based reinforcement learner on KGs \cite{PoLo}. Ratajczak et al. use expert-curated meta-paths to prune out information from a graph that experts would not consider relevant for specific learning tasks \cite{kg-filtering}.

\section{Handling Extreme Topological Variation}
\begin{figure}
\centering
\includegraphics[width=3in]{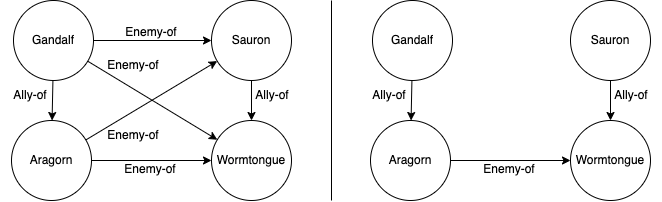}
\caption{Two graphs with distinct topology but identical information.} \label{fig2}
\end{figure}

The second issue facing KG learning systems to-date is that of their inability to handle extremities in node degree and topological variation. We formally define this problem as so: \textit{handling topological variation means capturing the semantics of a node or edge equally regardless of its local connectivity patterns}. An example of this can be seen in Figure~\ref{fig2}. Both graphs on the right contain identical semantic information. However, the topology of the graphs vary, which means that the degree of each node is greatly reduced in the right-hand graph as opposed to the left-hand one.

While the fact that Gandalf is an enemy of Sauron is implicit in the graph and obvious to a human, it is not labelled explicitly in the graph. As the remainder of this section will show, this means that traditional KG learning methods will be much less apt to learn that Gandalf and Sauron are enemies. Using this example as a guide, we formalise the problem of handling topological variation as so: \textit{By what methods is it possible to teach a KG learning system to robustly and consistently learn knowledge with distinct topology but identical semantics?} We call methods that attempt to distinguish such graphs "connectivity-tolerant methods".

\subsection{Connectivity-Tolerant Methods}
Topological imbalance in KGs has a variety of negative effects on learning using traditional KGE models, GNNs, and path-based KG learning systems \cite{kg-filtering,topological-imbalance,meta-learning-tail,meta-tail-gnn}. Low-degree nodes embed at a much lower quality relative to high-degree nodes \cite{meta-learning-tail}, and high-degree nodes are sometimes predicted as answers during inference simply because of their higher degree, not because of domain relevance \cite{topological-imbalance}. So-called "super-hubs", or nodes with extremely high degree in the graph, also dilute information and hinder learning; this occurs even when those nodes are highly relevant to the given domain \cite{kg-filtering,topological-imbalance}.

Existing solutions to topological imbalance include the work of Liu et al., which found that using meta-learning to make low-degree node embeddings more similar to high-degree node embeddings improved performance of KGEs on low-degree nodes \cite{meta-learning-tail}. The authors replicated this in GNNs with similar results \cite{meta-tail-gnn}. A study by Tang et al. found that explicitly modelling for node degree in the GNN allowed the model to more robustly learn embeddings for low-degree nodes \cite{sl-dsgcn}. However, even with these systems the core problem of poor representations for low-degree node remains at best mitigated, not solved, since modern KG learning paradigms rely principally on (a plurality of) local connections for learning \cite{sl-dsgcn,meta-learning-tail,meta-tail-gnn}.

\section{Focusing Learning}
The third issue facing KG learning systems is that, even though many KG learning-based tasks only have one type of prediction in mind, they attempt to learn all predictive tasks with equal strength \cite{kg-filtering}. Moreover, all triples in the input graph are generally considered equally relevant and true \cite{rml-review,kge-survey}. This means that KG learning models cannot estimate of how certain or relevant a fact is, nor use human-known uncertainty in how it models KG data. We formally define the problem of modelling uncertainty as so: \textit{focused learning is learning that explicitly models which triples are less certain or less relevant during training}.

\begin{figure}
\centering
\includegraphics[width=3in]{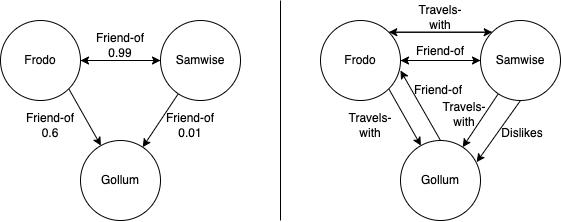}
\caption{Graph representing cases where we may care about uncertainty and relevance of various triples.} \label{fig3}
\end{figure}

We split this into two (somewhat overlapping) cases: dealing with uncertainty and dealing with relevance. For example, take Figure~\ref{fig3}. On the left, we have a graph where scores are given representing how uncertain (near 0) or certain (near 1) a fact is. On the right, another graph contains triples with no annotations, but where it is possible that some connections are uncertain. For instance, If Frodo is friends with Samwise, and Samwise dislikes Gollum, we may be less certain that Gollum is friends with Frodo. Similarly, we might not care about predicting friendship, but about predicting who travels with whom. This would make only the \textit{Travels-with} relationship directly relevant; other relationships would be of use only to the extend that they assist learning to predict triples about friendship.

Using this example as a guide, we formalise the problem of explicitly modelling for uncertainty as so: \textit{By what methods is it possible to distinguish certainty and relevance of triples within a KG?} We call methods that attempt to model uncertainty / relevance "focused learning methods".

\subsection{Focused Learning Methods}
One method for focused learning in the state-of-the-art is that taken by UKGE and FocusE for KGEs: to use uncertainty / relevance information given as additional triple-level labels in the KG, as is shown in the left of Figure~\ref{fig3} \cite{focuse,ukge}. Specifically, UKGE uses a set of out-of-band logical rules to model how uncertainties interact, and thus only works with uncertainty \cite{ukge}. FocusE takes a more broad approach: it directly uses numeric labels to modify the scoring layer of a KGE model with fewer assumptions, allowing it to model both uncertainty and relevance equally well \cite{focuse}.

Uncertainty can also be modelled implicitly, in the absence of certainty or relevance labels in the graph \cite{sl-dsgcn}. One method for this is used in SL-DSGCN, a GNN method that models uncertainty using a Bayesian-based teacher-student model \cite{sl-dsgcn}. The Bayesian Neural Network is trained not only to teach a student network, but to also give uncertainty scores as it teaches \cite{sl-dsgcn}. These scores are not input to the model, but learned and created automatically by the Bayesian Network during training \cite{sl-dsgcn}. 

Finally, using expert-curated meta-path information can also help focus learning on those specific paths \cite{PoLo,kg-filtering}. Two recent applications are filtering methods that use meta-paths to select for only task-relevant information as a pre-processing step \cite{kg-filtering} and the PoLo model, which uses reinforcement learning and logical rules to explicitly reward learning done on the given meta-paths.

\section{Explainability}
For our purposes, we define explainability in keeping with what Lipton calls "post-hoc interpretability", or \textit{the ability to explain why a model made a certain prediction} \cite{xai-defs}. Similarly, we say a method is explainable by design if it was created to natively give such explanations. Under this definition of explainability, almost all modern KGEs, GNNs, and path-based methods would not be explainable by design since their predictions do not provide any method for post-hoc explanation \cite{rml-review,kge-survey,rdf2vec,gnn-review}.

\begin{figure}
\centering
\includegraphics[width=3in]{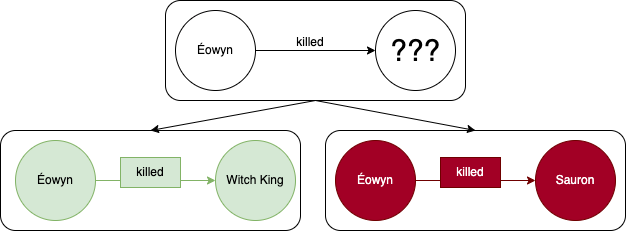}
\caption{A triple predicted to be true (light green) and false (dark red). Explanation aims to distinguish for what reasons one is labelled as true and the other as false.} \label{fig4}
\end{figure}

Suppose we have a trained KG learning system that performs link prediction. As illustrated in Figure~\ref{fig4}, it will attempt to predict if a triple is true (light green) or false (dark red); or in other words, distinguish why one graph structure is predicted to exist while the other is not. This leads us to formalise the problem of explanation of KG learning systems as so: \textit{By what methods can we extract the reason(s) that a KG learning system predicted the presence of a certain graph structure or feature rather than any other?} We call systems that implement these methods "Explainability Systems".

\subsection{Explainability Systems}
The first general approach to explainability is to provide a post-hoc explanation of a KG learning system that is not explainable by design, such as TransE \cite{rml-review}, RDF2vec \cite{rdf2vec}, or SL-DSGCN \cite{sl-dsgcn}. Methods for post-hoc interpretability of KGEs are generally based on estimating the influence of a triple on a given prediction \cite{interp-and-robustness-lp}. This includes various Instance Attribution Methods such as Influence Functions and Instance and Gradient Similarity computation \cite{kge-inst-attrr-methods}. For GNNs, many post-hoc explanation systems are not GNN-specific \cite{gnn-xai-exp-survey,gnn-xai-taxonomy}, although some, such as GNNExplainer, PGExplainer, and GraphMask explicitly take advantage of the triple-based structure of graphs to identify the most important triples for a prediction \cite{gnn-xai-exp-survey}. More detailed analyses can be found in \cite{interp-and-robustness-lp,gnn-xai-exp-survey,gnn-xai-taxonomy}.

The second approach is to create models that are explainable by design. This includes rule-based models such as AMIE+ and DL-Learner, which use symbolic logic to infer rules that are used to predict new statements in a graph \cite{amie,dl-learner}. Since those rules are written in human-readable logical clauses, they are direct explanations for the models' predictions. Models such as PoLo also fall in this category: it uses expert-curated rules (in the form of meta-paths) to create a policy-based model able to predict new triples \cite{PoLo}.

\section{Veni, Vidi, Vici: Towards a Common Conceptual Space}
\begin{figure}
\centering
\includegraphics[width=2in]{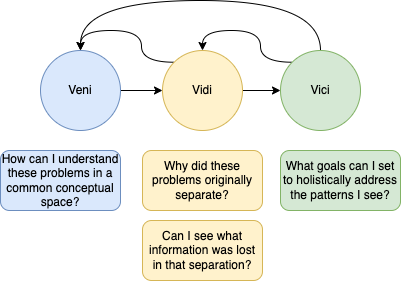}
\caption{An overview of the "Veni, Vidi, Vici" framework.} \label{fig5}
\end{figure}

In examining the four major challenges faced by modern KG learning systems, we observe that, while many papers attempt to solve each problem individually, none attempt to solve all four at once. The current state-of-the-art has divide-and-conquer methodology that results in tunnel-vision with respect to specific problems with KG learning systems and a lack of attention to the broader systemic shortcomings of the field as a whole. To this, we propose a new design methodology, called "Veni, Vidi, Vici" for helping human researchers to conceptualise, frame, and solve distinct problems as one. While we present this in the context of KG learning systems, it is potentially applicable to any case where multiple challenges exist within the same broader domain. An overview of this model and its use is shown in Figure~\ref{fig5}.

The "Veni, Vidi, Vici" approach occurs in three phases. The first, "Veni" (meaning "I came" in Latin) is to phrase all problems in a shared conceptual space. The second, "Vidi" (meaning "I saw" in Latin) is to deeply explore and analyse this shared conceptual space: for example, what information is lost when it splits into various sub-problems, and what elements of this space do human users care most about? The third and final, "Vici" (meaning "I conquered' in Latin) is to determine how to solve the underlying problem represented by this shared conceptual space as a single, atomic unit

As an example of the use of the "Veni, Vidi, Vici" system, consider the four challenges to KG learning systems presented here. The first step, Veni, we did by taking each of these challenges and placing them in the same shared conceptual space of distinguishing elements of graphs. Our Vidi step was to examine each of these problems in the state-of-the-art and their limitations, especially in terms of opportunity lost by treating each problem as distinct. Finally, our Vici step of creating, evaluating, and publishing a joint solution is left as a future direction -- one that this methodology well prepares up to take.

 We expand on these three phases below.

\subsection{Phase 1: Veni}
The Veni phase is characterised by asking, \textit{how?} Specifically, it asks \textit{How can I understand these problems in a shared conceptual space?} It aims to express each of the four issues -- expert knowledge integration, wide topological variation, attention to certainty and relevance, and explainable -- into the same terms.

The idea here is twofold. First, that considering the problems in isolation will lead to less general and less generalisable solutions to advancing KGE systems, as outlined in the review above. The second is that, since there is a single common goal (improving the predictive performance of KGE systems), that the various barriers to this goal should be expressible in common, unifying terms. The Veni step thus would not ask "To what extent can some KGE method reduce the detrimental effects of topological extremity in a KG?" -- such a question, while valuable, only focusses on one part of the problem, and does not take a fully holistic approach.

Instead, it begins by looking for common ground in these problems. The review above highlights that the methods that seek to strongly and meaningfully integrate expert knowledge for inference on a graph (such as Polo) tend to provide ways of estimating relevance and reducing the effect of extreme topological variation \cite{PoLo}. We also note that most logic-based methods tend to be more natively explainable than those that do not directly consider logic \cite{amie,dl-learner,rml-review,kge-survey}.

As a result, one possible question in the Veni phase would be the following: "To what extent can path-based models of graph logic represent the relevant semantic content of a KG?" Here, the common term chosen is logic -- most commonly supplied by an ontology for KGEs. It then asks how a logic based method can represent content that is \textit{relevant} and \textit{semantic}, meaning that it should be able to distinguish different semantics in identical structures, and focus learning on the most important parts thereof. In other words, this is all about  distinguishing graph elements, as outlined in the previous section -- distinguishing the meaning of regions with same structure (logic and semantics), distinguishing the relevance of distinct triples, and identifying the structures of a graph that actually encode similar information (logic and topological awareness). 

Once this core question is obtained, the Veni step is complete -- each problem has been presented in terms of the same basic building-blocks.

\subsection{Phase 2: Vidi}
The Vidi phase is characterised by asking \textit{why?}. Specifically, it asks two questions \textit{Why did these problems originally separate?} and \textit{How can I see what information was lost in that separation?}. The purpose of this phase is to solidify understanding of the question asked in the Veni phase and, critically, to ensure that the common terms being used to represent the problems in the same conceptual space actually address the core issue of why those problems should be modelled as one, rather than as individual units.

The review conducted here provides a hypothesis to answer these questions for KGEs: that these problems originally separated because of the distinct deficiencies identified by different researchers, and because of the differing aims and purposes for which various groups developed and used KGEs. Similarly, the main loss was that each of these problems was solved for a very specific purpose, and that therefore general advancement of KGEs was second to advancement in the specific areas most pressing at the time the research was conducted. In other words, there remains a large region of untapped potential for KGE development in providing more general and generalisable solutions to KG learning.

\subsection{Phase 3: Vici}
The Vici phase is characterised by asking \textit{what?} Specifically, it asks \textit{What goals can I set to holistically address the patterns I see?} This is the action step -- the step where the goals set are directly actionable. The output here is a specific model or set of models, evaluation protocols, and tests to ensure that any proposed solution not only answers the main question asked in the Veni phase, but that it actually improves on each of the distinct problems identified and recovers what was lost when each problem was taken separately (as identified in the Vidi phase).

\subsection{Bringing it all Together}
The purpose of the Veni, Vidi, Vici model is to invite a new paradigm of developing KGEs -- to focus on a holistic, broad goal rather than a variety of distinct sub-goals. Based on the existing literature and the review conducted here, it is in the opinion of the authors that a holistic approach to furthering KGE development will provide a novel, and potentially very successful, new method for approaching semantic modelling and KG representation.

It is critical to note that the proposed "Veni, Vidi, Vici" is not a linear model -- it is expected that each phase will involve multiple steps of backtracking to the previous (or first) stage for iterative refinement. It also is meant to be used in any case where several challenges exist within a single broader domain, and has potential to drive innovation in many fields outside of KG learning systems, although exploration of how it can be applied in other domains is left as a future direction.

\section{Conclusion}
In this paper, we provide a survey of the state-of-the-art in KG learning systems and the challenges they face. We conclude that the current "divide and conquer" approach to advancing individual problems facing KG learning systems is actively detrimental to advancing KG learning systems. To address this, we propose the "Veni, Vidi, Vici" system that allows human experts to reconsider these problems in a common framework. This framework explicitly models human desiderata and aims to help guide development along the directions that human experts desire most. It simultaneously aims to ensure co-empowerment not only of all areas of learning systems, but also how humans can interface with those systems. Finally, we note that "Veni, Vidi, Vici" is broadly applicable to any domain with multiple sub-problems.

\bibliographystyle{vancouver.bst}
\bibliography{vvv.bib}

\begin{thebibliography}{10}

\bibitem{kg-ovewview}
Hogan A, Blomqvist E, Cochez M, D’amato C, Melo GD, Gutierrez C, et~al.
\newblock Knowledge Graphs.
\newblock ACM Comput Surv. 2021 jul;54(4).
\newblock Available from: \url{https://doi.org/10.1145/3447772}.

\bibitem{hetionet}
Himmelstein DS, Lizee A, Hessler C, Brueggeman L, Chen SL, Hadley D, et~al.
\newblock Systematic integration of biomedical knowledge prioritizes drugs for
  repurposing.
\newblock eLife. 2017 sep;6:e26726.
\newblock Available from: \url{https://doi.org/10.7554/eLife.26726}.

\bibitem{light-into-the-dark}
Ali M, Berrendorf M, Hoyt CT, Vermue L, Galkin M, Sharifzadeh S, et~al.
\newblock Bringing Light Into the Dark: A Large-Scale Evaluation of Knowledge
  Graph Embedding Models Under a Unified Framework.
\newblock IEEE Transactions on Pattern Analysis and Machine Intelligence.
  2022;44(12):8825-45.

\bibitem{conceptnet}
Speer R, Chin J, Havasi C.
\newblock ConceptNet 5.5: An Open Multilingual Graph of General Knowledge.
\newblock In: Proceedings of the Thirty-First AAAI Conference on Artificial
  Intelligence. AAAI'17. AAAI Press; 2017. p. 4444–4451.

\bibitem{rml-review}
Nickel M, Murphy K, Tresp V, Gabrilovich E.
\newblock A Review of Relational Machine Learning for Knowledge Graphs.
\newblock Proceedings of the IEEE. 2016;104(1):11-33.

\bibitem{kge-survey}
Wang Q, Mao Z, Wang B, Guo L.
\newblock Knowledge Graph Embedding: A Survey of Approaches and Applications.
\newblock IEEE Transactions on Knowledge and Data Engineering.
  2017;29(12):2724-43.

\bibitem{PoLo}
Liu Y, Hildebrandt M, Joblin M, Ringsquandl M, Raissouni R, Tresp V.
\newblock Neural Multi-Hop Reasoning With Logical Rules on Biomedical Knowledge
  Graphs.
\newblock In: Extended Semantic Web Conference; 2021. .

\bibitem{dl-learner}
Bühmann L, Lehmann J, Westphal P.
\newblock DL-Learner—A framework for inductive learning on the Semantic Web.
\newblock Journal of Web Semantics. 2016;39:15-24.
\newblock Available from:
  \url{https://www.sciencedirect.com/science/article/pii/S157082681630018X}.

\bibitem{kge-logic-survey}
Zhang W, Chen J, Li J, Xu Z, Pan J, Chen H. Knowledge Graph Reasoning with
  Logics and Embeddings: Survey and Perspective; 2022.

\bibitem{rdf2vec}
Ristoski P, Rosati J, Noia TD, Leone RD, Paulheim H.
\newblock RDF2Vec: RDF graph embeddings and their applications.
\newblock Semantic Web. 2019;10:721-52.

\bibitem{sl-dsgcn}
Tang X, Yao H, Sun Y, Wang Y, Tang J, Aggarwal C, et~al.
\newblock Investigating and Mitigating Degree-Related Biases in Graph
  Convoltuional Networks.
\newblock In: Proceedings of the 29th ACM International Conference on
  Information \& Knowledge Management. CIKM '20. New York, NY, USA: Association
  for Computing Machinery; 2020. p. 1435–1444.
\newblock Available from: \url{https://doi.org/10.1145/3340531.3411872}.

\bibitem{gnn-review}
Zhou J, Cui G, Hu S, Zhang Z, Yang C, Liu Z, et~al.
\newblock Graph neural networks: A review of methods and applications.
\newblock AI Open. 2020;1:57-81.
\newblock Available from:
  \url{https://www.sciencedirect.com/science/article/pii/S2666651021000012}.

\bibitem{vgae}
Kipf T, Welling M.
\newblock Variational Graph Auto-Encoders.
\newblock ArXiv. 2016;abs/1611.07308.

\bibitem{CensNet}
Jiang X, Zhu R, Li S, Ji P.
\newblock Co-embedding of Nodes and Edges with Graph Neural Networks.
\newblock IEEE Transactions on Pattern Analysis and Machine Intelligence.
  2020:1-1.

\bibitem{kg-filtering}
Ratajczak F, Joblin M, Ringsquandl M, Hildebrandt M.
\newblock Task-driven knowledge graph filtering improves prioritizing drugs for
  repurposing.
\newblock BMC bioinformatics. 2022 March;23(1):84.
\newblock Available from: \url{https://europepmc.org/articles/PMC8894843}.

\bibitem{kges-for-ddi}
Mohamed SK, Nounu A, Nováček V.
\newblock {Biological applications of knowledge graph embedding models}.
\newblock Briefings in Bioinformatics. 2020 02;22(2):1679-93.
\newblock Available from: \url{https://doi.org/10.1093/bib/bbaa012}.

\bibitem{kges-for-ddi-2}
Celebi R, Uyar H, Yasar E, Gümüs O, Dikenelli O, Dumontier M.
\newblock Evaluation of knowledge graph embedding approaches for drug-drug
  interaction prediction in realistic settings.
\newblock BMC Bioinformatics. 2019 12;20.

\bibitem{onto-in-neg-sampler}
Jain N, Tran TK, Gad-Elrab M, Stepanova D.
\newblock In: Improving Knowledge Graph Embeddings with Ontological Reasoning;
  2021. p. 410-26.

\bibitem{realistic-ddi}
Celebi R, Uyar H, Yasar E, Gumus O, Dikenelli O, Dumontier M.
\newblock Evaluation of knowledge graph embedding approaches for drug-drug
  interaction prediction in realistic settings.
\newblock BMC bioinformatics. 2019;20(1):1-14.

\bibitem{kge-rare-disease}
Sosa DN, Derry A, Guo M, Wei E, Brinton C, Altman RB.
\newblock A literature-based knowledge graph embedding method for identifying
  drug repurposing opportunities in rare diseases.
\newblock In: Pacific Symposium on Biocomputing 2020. World Scientific; 2019.
  p. 463-74.

\bibitem{discover-protein-targets}
Mohamed SK, Nov{\'a}{\v{c}}ek V, Nounu A.
\newblock Discovering protein drug targets using knowledge graph embeddings.
\newblock Bioinformatics. 2020;36(2):603-10.

\bibitem{target-drug-discovery}
Mohamed SK, Nounu A, Nov\'{a}\v{c}ek V.
\newblock Drug Target Discovery Using Knowledge Graph Embeddings.
\newblock In: Proceedings of the 34th ACM/SIGAPP Symposium on Applied
  Computing. SAC '19. New York, NY, USA: Association for Computing Machinery;
  2019. p. 11–18.
\newblock Available from: \url{https://doi.org/10.1145/3297280.3297282}.

\bibitem{topological-imbalance}
Bonner S, Kirik U, Engkvist O, Tang J, Barrett IP.
\newblock {Implications of topological imbalance for representation learning on
  biomedical knowledge graphs}.
\newblock Briefings in Bioinformatics. 2022 07;23(5).
\newblock Bbac279.
\newblock Available from: \url{https://doi.org/10.1093/bib/bbac279}.

\bibitem{meta-learning-tail}
Liu Z, Zhang W, Fang Y, Zhang X, Hoi SCH.
\newblock Towards Locality-Aware Meta-Learning of Tail Node Embeddings on
  Networks.
\newblock In: Proceedings of the 29th ACM International Conference on
  Information \& Knowledge Management. CIKM '20. New York, NY, USA: Association
  for Computing Machinery; 2020. p. 975–984.
\newblock Available from: \url{https://doi.org/10.1145/3340531.3411910}.

\bibitem{meta-tail-gnn}
Liu Z, Nguyen TK, Fang Y.
\newblock Tail-GNN: Tail-Node Graph Neural Networks.
\newblock In: Proceedings of the 27th ACM SIGKDD Conference on Knowledge
  Discovery \& Data Mining. KDD '21. New York, NY, USA: Association for
  Computing Machinery; 2021. p. 1109–1119.
\newblock Available from: \url{https://doi.org/10.1145/3447548.3467276}.

\bibitem{focuse}
Pai S, Costabello L.
\newblock Learning Embeddings from Knowledge Graphs With Numeric Edge
  Attributes.
\newblock In: Zhou ZH, editor. Proceedings of the Thirtieth International Joint
  Conference on Artificial Intelligence, {IJCAI-21}. International Joint
  Conferences on Artificial Intelligence Organization; 2021. p. 2869-75.
\newblock Main Track.
\newblock Available from: \url{https://doi.org/10.24963/ijcai.2021/395}.

\bibitem{explaining-kges}
Bianchi F, Rossiello G, Costabello L, Palmonari M, Minervini P. Knowledge Graph
  Embeddings and Explainable AI; 2020.

\bibitem{amie}
Gal\'{a}rraga L, Teflioudi C, Hose K, Suchanek FM.
\newblock Fast Rule Mining in Ontological Knowledge Bases with AMIE+.
\newblock The VLDB Journal. 2015 dec;24(6):707–730.
\newblock Available from: \url{https://doi.org/10.1007/s00778-015-0394-1}.

\bibitem{xai-metrics}
Doshi-Velez F, Kim B.
\newblock Towards A Rigorous Science of Interpretable Machine Learning.
\newblock arXiv: Machine Learning. 2017.

\bibitem{itere}
Zhang W, Paudel B, Wang L, Chen J, Zhu H, Zhang W, et~al.
\newblock Iteratively Learning Embeddings and Rules for Knowledge Graph
  Reasoning.
\newblock In: The World Wide Web Conference. WWW '19. New York, NY, USA:
  Association for Computing Machinery; 2019. p. 2366–2377.
\newblock Available from: \url{https://doi.org/10.1145/3308558.3313612}.

\bibitem{Query2box}
Ren* H, Hu* W, Leskovec J.
\newblock Query2box: Reasoning over Knowledge Graphs in Vector Space Using Box
  Embeddings.
\newblock In: International Conference on Learning Representations; 2020.
  Available from: \url{https://openreview.net/forum?id=BJgr4kSFDS}.

\bibitem{cqd}
Minervini P, Arakelyan E, Daza D, Cochez M.
\newblock Complex Query Answering with Neural Link Predictors (Extended
  Abstract)*.
\newblock In: Raedt LD, editor. Proceedings of the Thirty-First International
  Joint Conference on Artificial Intelligence, {IJCAI-22}. International Joint
  Conferences on Artificial Intelligence Organization; 2022. p. 5309-13.
\newblock Sister Conferences Best Papers.
\newblock Available from: \url{https://doi.org/10.24963/ijcai.2022/741}.

\bibitem{deductive-completion}
Alshahrani M, Khan MA, Maddouri O, Kinjo AR, Queralt-Rosinach N, Hoehndorf R.
\newblock {Neuro-symbolic representation learning on biological knowledge
  graphs}.
\newblock Bioinformatics. 2017 04;33(17):2723-30.
\newblock Available from: \url{https://doi.org/10.1093/bioinformatics/btx275}.

\bibitem{e2r}
Garg D, Ikbal S, Srivastava SK, Vishwakarma H, Karanam H, Subramaniam LV.
\newblock Quantum Embedding of Knowledge for Reasoning.
\newblock In: Wallach H, Larochelle H, Beygelzimer A, d\textquotesingle
  Alch\'{e}-Buc F, Fox E, Garnett R, editors. Advances in Neural Information
  Processing Systems. vol.~32. Curran Associates, Inc.; 2019. Available from:
  \url{https://proceedings.neurips.cc/paper/2019/file/cb12d7f933e7d102c52231bf62b8a678-Paper.pdf}.

\bibitem{el-embeddings}
Kulmanov M, Liu-Wei W, Yan Y, Hoehndorf R.
\newblock El embeddings: Geometric construction of models for the description
  logic el++.
\newblock arXiv preprint arXiv:190210499. 2019.

\bibitem{ukge}
Chen X, Chen M, Shi W, Sun Y, Zaniolo C.
\newblock Embedding Uncertain Knowledge Graphs.
\newblock In: Proceedings of the Thirty-Third AAAI Conference on Artificial
  Intelligence and Thirty-First Innovative Applications of Artificial
  Intelligence Conference and Ninth AAAI Symposium on Educational Advances in
  Artificial Intelligence. AAAI'19/IAAI'19/EAAI'19. AAAI Press; 2019. Available
  from: \url{https://doi.org/10.1609/aaai.v33i01.33013363}.

\bibitem{xai-defs}
Lipton ZC.
\newblock The Mythos of Model Interpretability.
\newblock Commun ACM. 2018 sep;61(10):36–43.
\newblock Available from: \url{https://doi.org/10.1145/3233231}.

\bibitem{interp-and-robustness-lp}
Pezeshkpour P, Tian Y, Singh S.
\newblock Investigating Robustness and Interpretability of Link Prediction via
  Adversarial Modifications.
\newblock In: Proceedings of the 2019 Conference of the North {A}merican
  Chapter of the Association for Computational Linguistics: Human Language
  Technologies, Volume 1 (Long and Short Papers). Minneapolis, Minnesota:
  Association for Computational Linguistics; 2019. p. 3336-47.
\newblock Available from: \url{https://aclanthology.org/N19-1337}.

\bibitem{kge-inst-attrr-methods}
Bhardwaj P, Kelleher J, Costabello L, O{'}Sullivan D.
\newblock Adversarial Attacks on Knowledge Graph Embeddings via Instance
  Attribution Methods.
\newblock In: Proceedings of the 2021 Conference on Empirical Methods in
  Natural Language Processing. Online and Punta Cana, Dominican Republic:
  Association for Computational Linguistics; 2021. p. 8225-39.
\newblock Available from: \url{https://aclanthology.org/2021.emnlp-main.648}.

\bibitem{gnn-xai-exp-survey}
Li P, Yang Y, Pagnucco M, Song Y. Explainability in Graph Neural Networks: An
  Experimental Survey; 2022.

\bibitem{gnn-xai-taxonomy}
Yuan H, Yu H, Gui S, Ji S. Explainability in Graph Neural Networks: A Taxonomic
  Survey; 2020.

\end{thebibliography}

\end{document}